\documentclass[letterpaper, 10 pt, journal, twoside]{IEEEtran}
\usepackage{booktabs}
\usepackage{multirow}
\usepackage{graphicx}
\usepackage{makecell}
\usepackage{amsmath}
\usepackage{amssymb}
\usepackage{rotating}
\usepackage{threeparttable}
\usepackage{hyperref}
\usepackage{xcolor}
\usepackage{ulem}
\ifCLASSINFOpdf
\else
\fi
\begin{document}
%
\title{GICI-LIB: A GNSS/INS/Camera Integrated Navigation Library}
%
%
%

\author{Cheng Chi, Xin Zhang, \textit{Member, IEEE}, Jiahui Liu, Yulong Sun, Zihao Zhang, \\ and Xingqun Zhan, \textit{Senior Member, IEEE} 
\thanks{Manuscript received: June 21, 2023; Revised: September 17, 2023; Accepted: October 9, 2023. This paper was recommended for publication by Editor Sven Behnke upon evaluation of the Associate Editor and Reviewers’ comments. This work was supported by National Key R\&D Program of China (2022YFB3904401). (\textit{Corresponding author: Xin Zhang})}
\thanks{The authors are with School of Aeronautics and Astronautics,
        Shanghai Jiao Tong University, Shanghai, China. (e-mail: \url{chichengcn@sjtu.edu.cn}; \url{xin.zhang@sjtu.edu.cn}; \url{jh.liu@sjtu.edu.cn}; \url{sunyulong@sjtu.edu.cn}; \url{zh.zhang@sjtu.edu.cn}; \url{xqzhan@sjtu.edu.cn}) }
\thanks{The source code, documentation, and dataset are available at 
\textcolor{blue}{\uline{\url{https://github.com/chichengcn/gici-open}}}}
\thanks{Digital Object Identifier (DOI): see top of this page.}
}
%
%

\markboth{IEEE Robotics and Automation Letters. Preprint Version. Accepted October, 2023}
{Chi \MakeLowercase{\textit{et al.}}: GICI-LIB: A GNSS/INS/Camera Integrated Navigation Library} 

%



\maketitle

\begin{abstract}
Accurate navigation is essential for autonomous robots and vehicles. In recent years, the integration of the Global Navigation Satellite System (GNSS), Inertial Navigation System (INS), and camera has garnered considerable attention due to its robustness and high accuracy in diverse environments. However, leveraging the full capacity of GNSS is cumbersome because of the diverse choices of formulations, error models, satellite constellations, signal frequencies, and service types, which lead to different precision, robustness, and usage dependencies. To clarify the capacity of GNSS algorithms and accelerate the development efficiency of employing GNSS in multi-sensor fusion algorithms, we open source the GNSS/INS/Camera Integration Library (GICI-LIB), together with detailed documentation and a comprehensive land vehicle dataset. A factor graph optimization-based multi-sensor fusion framework is established, which combines almost all GNSS measurement error sources by fully considering temporal and spatial correlations between measurements. The graph structure is designed for flexibility, making it easy to form any kind of integration algorithm. For illustration, Real-Time Kinematic (RTK), Precise Point Positioning (PPP), and four RTK-based algorithms from GICI-LIB are evaluated using our dataset and public datasets. Results confirm the potential of the GICI system to provide continuous precise navigation solutions in a wide spectrum of urban environments. 
\end{abstract}


%
\IEEEpeerreviewmaketitle

\section{Introduction}

With the significant increase in computing capability and the simultaneous reduction in power consumption and cost, autonomous robots and vehicles have become ubiquitous \cite{huang2019visual}. For these applications, robust and accurate navigation is essential. Although the Visual-Inertial Navigation System (VINS) has been recognized as a practical solution due to its high accuracy and low cost over the years \cite{cadena2016past}, it still suffers from performance fluctuations under complex environments and pose drift during long-term cases \cite{janai2020computer}.

Fusing VINS and GINS together, namely GVINS, is an effective approach to substantially enhance the robustness and availability \cite{cao2022gvins}. The two systems are complementary. GINS can provide robust and globally accurate pose solutions in most outdoor settings and VINS can greatly constrain the pose drift rate under GNSS-challenging conditions. There are four typical GNSS formulations: Single Point Positioning (SPP), Real-Time Differential (RTD), Real-Time Kinematic (RTK), and Precise Point Positioning (PPP). SPP and RTD are meter-level accuracy formulations, while RTK and PPP are centimeter-level accuracy formulations. Previously, centimeter-level GNSS was commonly used in cost-insensitive applications, such as surveying and aviation. With the cost reduction of high-precision GNSS chips and augmentation services, centimeter-level GNSS devices are gaining momentum in autonomous applications \cite{whitepaper2022}. By incorporating high-precision GNSS positioning formulations, the GVINS system will be able to provide continuous centimeter-level solutions under various environments.

Constructing a GVINS system is not difficult, which has been done by several works \cite{cao2022gvins, niu2022ic, liu2021optimization, li2022p3vins}. However, fully utilizing the role of GNSS is cumbersome. Rigorously, there are multiple measurement formulations, error models, satellite constellations, signal frequencies, and service types. These distinct options can significantly impact precision, robustness, and usage dependencies. One can achieve an imprecise and vulnerable solution via SPP formulations, and can also achieve a precise and robust solution even in a typical urban environment via multi-constellations multi-frequency RTK with the augmentation of Observation Space Representation (OSR) services. In environments where OSR service is unavailable because of the absence of reference stations, one can still achieve a precise solution within a few minutes via PPP with the augmentation of State Space Representation (SSR) services.

To clarify the capacity of GNSS algorithms and accelerate the development efficiency of employing GNSS in multi-sensor fusion algorithms, we present the GNSS/INS/Camera Integration Library (GICI-LIB), together with detailed documentation and a comprehensive land vehicle dataset. The key features are highlighted as follows: 

\textbullet \ A factor graph optimization-based multi-sensor fusion framework, which implements most of the possible GNSS loose and tight integration factors, INS factors, visual factors, and motion constraints, together with reliable initialization, measurement sparsification, and outlier rejection algorithms. The GNSS formulations are implemented towards four constellations and full frequencies.

\textbullet \ Unlike other state-of-the-art (SOTA) integrated INS/GNSS/camera systems, our system focuses on combining almost all measurement error sources of GNSS in space, propagation, and ground segments by fully considering temporal and spatial correlations between measurements. It is in this way that our system outperforms the SOTA methods on the majority of our open-source datasets and the evaluated INS/GNSS/camera code repositories.

\textbullet \ For ease of use, the software is developed under object-oriented programming features, and the graph is designed to enable flexible addition of sensors. By simple instantiation, one can easily form any kind of multi-sensor fusion algorithm with considerable robustness. 

\textbullet \ An open land vehicle dataset that collects necessary raw data for all the above algorithms. The datasets contain multiple short-term and long-term trajectories, covering open-sky, tree-lined, typical urban, and dense urban environments. This is the first dataset that includes real-time OSR and SSR corrections accompanying with the raw GNSS observations.

The rest of this letter is structured as follows. In section \ref{sec:RelatedWork}, we review existing relevant works. Section \ref{sec:Overview} shows the structure of the proposed system. Section \ref{sec:Methodology} illustrates the formulation and methodology. In section \ref{sec:Experiments}, we evaluate our algorithms against other open-source algorithms. Finally section \ref{sec:Conclusion} concludes this letter.

\section{Related Work}
\label{sec:RelatedWork}

\begin{table}[htbp]
\centering
\begin{threeparttable}[b]
\caption{Summary of the most relevant optimization-based estimators for GNSS.} 
\label{tab:related_works}
\setlength{\tabcolsep}{1mm}
\renewcommand\arraystretch{1.2}
\belowrulesep=0pt
\aboverulesep=0pt
\begin{tabular}{c|c|c|c|c|c|c|c|c|c|c|c|c|c|c} 
    \toprule 
    {} & \multicolumn{2}{c|}{Type} & \multicolumn{2}{c|}{\makecell[c]{LC \\ cap.}} & \multicolumn{7}{c|}{\makecell[c]{TC \\ cap.}} & \multicolumn{2}{c|}{Perf.} & {} \\
    \hline
    {} & \rotatebox{90}{Sensors} & \rotatebox{90}{Fusion level} & \rotatebox{90}{Position} & \rotatebox{90}{Velocity} & \rotatebox{90}{Pseudorange} & \rotatebox{90}{Doppler} & \rotatebox{90}{Carrier phase} & \rotatebox{90}{OSR} & \rotatebox{90}{SSR} & \rotatebox{90}{Frequencies} & \rotatebox{90}{Constellations} & \rotatebox{90}{Precision level} & \rotatebox{90}{Robustness} & \rotatebox{90}{Open source}  \\
    \hline
    {\cite{sunderhauf2012towards}} & {G} & {-} & {-} & {-} & {$\checkmark$} & {$\times$} & {$\times$} & {$\times$} & {$\times$} & {S} & {G} & {m} & {} & {$\times$} \\
    \hline
    {\makecell[c]{Graph \\ GNSSLib \\ \cite{wen2021towards}}} & {G} & {-} & {-} & {-} & {$\checkmark$} & {$\checkmark$} & {$\checkmark$\tnote{1}} & {$\checkmark$\tnote{1}} & {$\times$} & {S} & {GC} & {dm} & { } & {$\checkmark$} \\
    \hline
    {\cite{gao2022robust}} & {G} & {-} & {-} & {-} & {$\checkmark$} & {$\checkmark$} & {$\checkmark$} & {$\checkmark$} & {$\times$} & {D} & {GC} & {cm} & {} & {$\times$} \\
    \hline
    {\makecell[c]{OB-GINS \\ \cite{tang2022impact}}} & {GI} & {LC} & {$\checkmark$} & {$\times$} & {-} & {-} & {-} & {-} & {-} & {-} & {-} & {cm\tnote{2}} & {F} & {$\checkmark$} \\
    \hline
    {\cite{li2018robust}} & {GI} & {TC} & {-} & {-} & {$\checkmark$} & {$\times$} & {$\times$} & {$\times$} & {$\times$} & {\tnote{3}} & {G} & {m} & {} & {$\times$} \\
    \hline
    {\cite{wen2021factor}} & {GI} & {\makecell[c]{LC \\ TC}} & {$\checkmark$} & {$\times$} & {$\checkmark$} & {$\times$} & {$\times$} & {$\times$} & {$\times$} & {S} & {GC} & {m} & {} & {$\times$} \\    
    \hline
    {\makecell[c]{VINS- \\ Fusion \\ \cite{qin2019general}}} & {GIC} & {SS} & {$\checkmark$} & {$\times$} & {-} & {-} & {-} & {-} & {-} & {-} & {-} & {cm} & {F} & {$\checkmark$} \\
    \hline
    {\makecell[c]{GOMSF \\ \cite{mascaro2018gomsf}}} & {GIC} & {SS} & {$\checkmark$} & {$\times$} & {-} & {-} & {-} & {-} & {-} & {-} & {-} & {dm} & {} & {$\times$} \\
    \hline
    {\makecell[c]{IC- \\ GVINS \\ \cite{niu2022ic}}} & {GIC} & {SRR} & {$\checkmark$} & {$\times$} & {-} & {-} & {-} & {-} & {-} & {-} & {-} & {cm} & {G} & {$\checkmark$} \\
    \hline
    {\cite{liu2021optimization}} & {GIC} & {RRR} & {-} & {-} & {$\checkmark$} & {$\checkmark$} & {$\times$} & {$\times$} & {$\times$} & {} & {GR} & {m} & {} & {$\times$} \\
    \hline
    {\makecell[c]{GVINS \\ \cite{cao2022gvins}}} & {GIC} & {RRR} & {-} & {-} & {$\checkmark$} & {$\checkmark$} & {$\times$} & {$\times$} & {$\times$} & {S} & {\makecell[c]{GR \\ EC}} & {m} & { } & {$\checkmark$} \\
    \hline
    {\makecell[c]{P3-VINS \\ \cite{li2022p3vins}}} & {GIC} & {RRR} & {-} & {-} & {$\checkmark$} & {$\checkmark$} & {$\checkmark$} & {$\times$} & {I} & {D} & {\makecell[c]{GR \\ EC}} & {dm} & {} & {$\times$} \\
    \hline
    {\makecell[c]{\textbf{GICI-} \\ \textbf{LIB}}} & {\makecell[c]{G \\ GI \\ GIC}} & {\makecell[c]{LC \\ TC \\ SRR \\ RRR}} & {$\checkmark$} & {$\checkmark$} & {$\checkmark$} & {$\checkmark$} & {$\checkmark$} & {$\checkmark$} & {\makecell[c]{I \\ II \\ III\tnote{4}}} & {M} & {\makecell[c]{GR \\ EC}} & {\makecell[c]{m \\ dm \\ cm}} & {E} & {$\checkmark$} \\
    \bottomrule 
\end{tabular} 
\begin{tablenotes}
 \item[1] Its RTK module was only designed and tested for static motion and is not capable of performing correctly under dynamic motion.
 \item[2] The precision level of GNSS loose integrations is influenced by the type of GNSS solution employed. We present the precision reported in the study here.
 \item[3] Blank means we cannot get the corresponding information.
 \item[4] We implemented the SSR-II and SSR-III features but did not instantiate them, because there is currently no standard and stable open real-time service that can be used.
\end{tablenotes}
\end{threeparttable}
\vspace{-1.5em}
\end{table}

In the past decades, almost all GNSS algorithms have utilized filter-based methods. Unlike GNSS, the state estimation problem for visual navigation can be categorized as either filter-based methods \cite{mourikis2007multi,li2013high} or optimization-based methods \cite{leutenegger2015keyframe, forster2016svo, qin2018vins}. Filter-based methods can achieve considerable accuracy while ensuring high computational efficiency. However, these methods are more  susceptible to linearization errors that have the potential to negatively impact the accuracy and robustness of the estimator \cite{huang2019visual}. In contrast, optimization-based methods excel at handling nonlinearity and achieve higher accuracy \cite{niu2022ic}, albeit with a trade-off in computational efficiency.

There are several optimization-based GNSS or GNSS integration algorithms reported. To clarify their usage of GNSS capacities, we summarize them together with our algorithms in Table \ref{tab:related_works}. The fusion types are categorized into GNSS-only (G), GNSS/INS (GI) Loose Couple (LC), GI Tight couple (TC), GNSS/INS/Camera (GIC) SS (Fusing GNSS solution with VINS solution), GIC Solution/Raw/Raw (SRR), and GIC Raw/Raw/Raw (RRR). The frequencies are categorized into Single (S), Double (D), and Multiple (M). The constellations are categorized into GPS (G), GLONASS (R), Galileo (E), and BDS (C). The robustness of the system is classified into three categories: Fair (F), Good (G), and Extraordinary (E). These categories are assigned based on the performance evaluation conducted in Section \ref{sec:Experiments}. There are three levels of SSR services: SSR-I contains precise ephemeris and code biases, SSR-II contains phase biases, and SSR-III contains local atmosphere delays, which enables PPP, PPP with ambiguity resolution (PPP-AR), and PPP with local reference network augmentation (PPP-RTK), respectively. We use the term PPP to collectively represent the three kinds of algorithms in GICI-LIB. 

It is clear that GICI-LIB fills lots of gaps for the optimization-based GNSS algorithms. It is the first optimization-based algorithm that utilizes GNSS velocity for LC, incorporating multi-frequency processing for TC, and supports multiple GNSS algorithms for multiple levels of integrations. These features effectively unlock the full potential of GNSS technology. Moreover, GICI-LIB is the first open-source software that offers full capacity of optimization-based RTK, PPP, and their tight integration with IMU and cameras. These modules are implemented with the concept of expansibility, greatly accelerating the development efficiency of employing GNSS in multi-sensor fusion algorithms. Except for the indexes listed in the table, our system also has various progresses on GNSS error handling, ambiguity resolution, outlier rejection, temporal sparsification, and initialization. These details bring our system extraordinary performance.

\section{System Overview}
\label{sec:Overview}

\begin{figure} [t]
    \vspace{1em}
    \centering
    \includegraphics[width=0.48\textwidth]{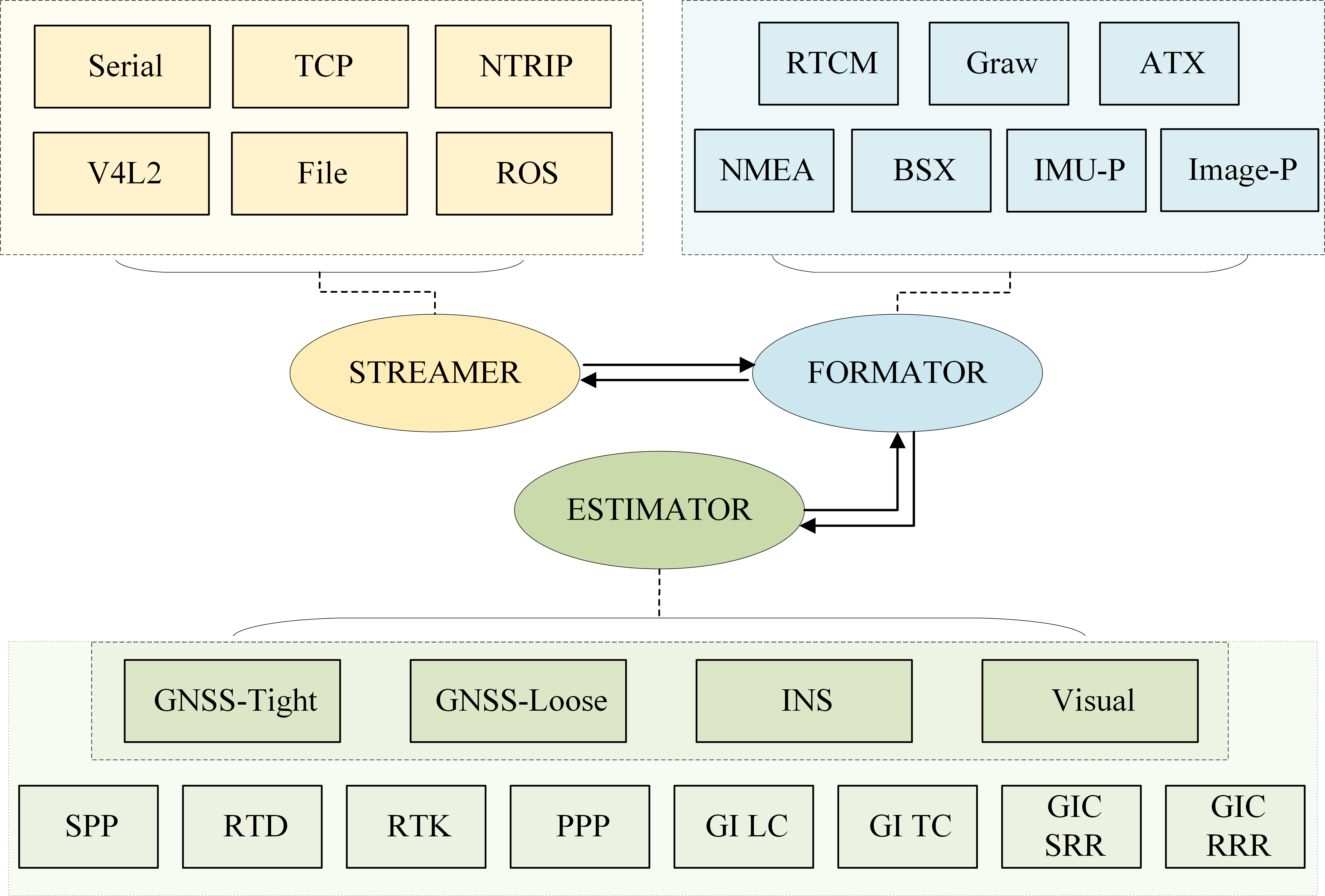}
    \caption{Structure of GICI-LIB.}
    \label{fig:system_structure}
\end{figure}

GICI-LIB is comprised of various hardware I/O controllers, data decoders, encoders, and estimation processors. These nodes operate in multiple threads, ensuring concurrent execution. The system structure is shown in Figure \ref{fig:system_structure}. By configuring nodes defined by our configuration file, the system can be used not only for real-time and post-processed algorithmic processing, but also for stream transformation and conversion. 

The estimator node is designed using a two-layer inheritance. The first layer implements necessary functionalities for each sensor, while the second layer inherits the sensor features to facilitate specific integration algorithms. The estimator node operates in its own dedicated thread, which is composed of three sub-threads: frontend, backend, and export. The frontend thread handles time-consuming preprocessing tasks. When fusing  GNSS, INS, and camera, the front-end thread will solely process the raw data from the camera since only the feature detection and tracking procedures are time-consuming. The backend conducts initialization and optimization with the Factor Graph Optimization (FGO) algorithm. Its solution can be used by the frontend to help improve the quality of feature tracking. Finally, the export thread will integrate the data from the backend thread according to desired timestamps. The integrated solution can be delivered to the stream threads for hardware output or file storage. Additionally, it can also be fed into other estimator threads for loose coupling.

As examples, in the code repository, we instantiate several algorithms for the estimator nodes, including SPP, RTD, RTK, PPP, SPP-based LC and TC GINS, SPP-based Solution/Raw/Raw SRR and RRR GVINS, RTK-based LC GINS, TC GINS, SSR GVINS, and RRR GVINS. 

\section{Methodology}
\label{sec:Methodology}

In order to emphasize the key points of this letter, we focus solely on the architecture design and GNSS positioning algorithms in this section. Details of additional features, such as the visual and INS factors, outlier detection algorithms, and measurement sparsification algorithms, will be omitted. Readers are invited to refer to our documentation for these omitted features.

In this section, we first introduce the FGO architecture. Afterward, we present GNSS factors, RTK algorithm, and PPP algorithm. Finally, we illustrate the GNSS ambiguity resolution process, which is the key technology for achieving centimeter-level positioning accuracy. 

\subsection{Factor Graph Optimization}
\label{sec:Methodology:FGO}

\begin{figure} [t]
    \vspace{1em}
    \centering
    \includegraphics[width=0.48\textwidth]{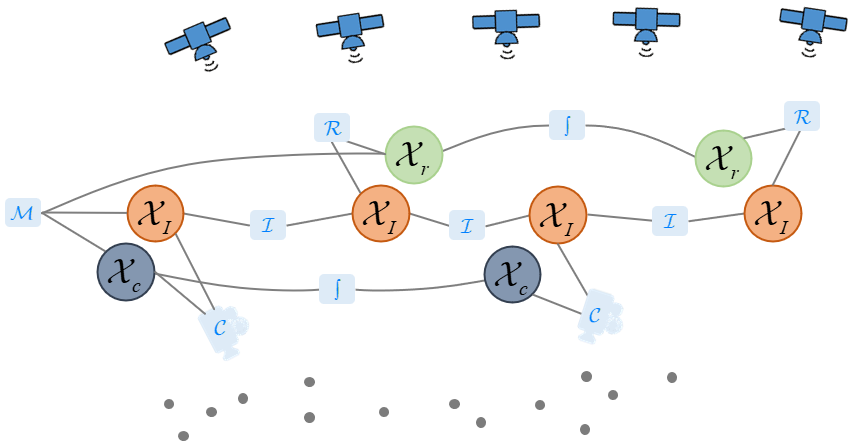}
    \caption{Structure of the FGO-based estimation. $\chi_r$, $\chi_I$, and $\chi_c$ represent the estimated parameters of the GNSS receiver, INS, and camera respectively. $\mathcal{M}$, $\mathcal{R}$, $\mathcal{C}$, $\mathcal{I}$, and $\int$ stands for the marginalization factor, the GNSS factors, the visual factors, the INS pre-integration factor, and the time-propagation factors, respectively.}
    \label{fig:fgo_structure}
\end{figure}

The GVINS integration structure is shown in Figure \ref{fig:fgo_structure}. The GNSS-only and GNSS/INS integration graphs are subsets of the GVINS integration graph, and thus, are not individually delineated here. This graph describes a non-linear Least-Square (LSQ) problem
\begin{align}
\hat{\boldsymbol{\chi}} = \mathop{\arg\min}\limits_{\boldsymbol{\chi}}
& \{ \|\boldsymbol{z}_p - \boldsymbol{H}_p \boldsymbol{\chi}\|^2 + \|\boldsymbol{z}_r - h_r(\boldsymbol{\chi}_r, \boldsymbol{\chi}_I)\|^2 + \notag \\
& \|\boldsymbol{z}_I - h_I(\boldsymbol{\chi}_I)\|^2 + \|\boldsymbol{z}_c - h_c(\boldsymbol{\chi}_c, \boldsymbol{\chi}_I)\|^2 \}
\label{eq:lsq}
\end{align}
where $\boldsymbol{\chi} = [\boldsymbol{\chi}_I, \boldsymbol{\chi}_r, \boldsymbol{\chi}_c]$ are the parameters to be estimated. The subscripts $r$, $I$, and $c$ represent GNSS receiver, INS, and camera, respectively. $\boldsymbol{z}$ is the measurements. $h$ is the corresponding non-linear measurement models. The constitution of the parameters and measurements varies depending on the utilized formulation. We will introduce them later. $\boldsymbol{z}_p$ and $\boldsymbol{H}_p$ are the pseudo-measurement and linearized Jacobian of prior information, which are computed during marginalization.

\begin{figure} [t]
    \vspace{1em}
    \centering
    \includegraphics[width=0.48\textwidth]{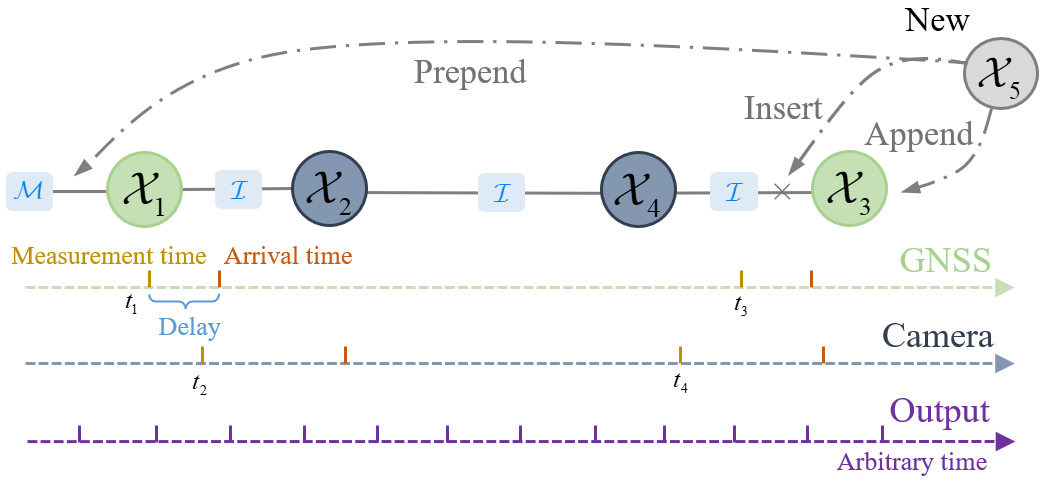}
    \caption{Flexibility of the FGO structure. }
    \label{fig:fgo_addition}
\end{figure}

The graph is designed towards flexibility, see Figure \ref{fig:fgo_addition}. The flexibility is enabled by three features: 1) The measurement timestamps of different sensors are not necessarily aligned. 2) New measurement data can be added to the graph at any point, not necessarily at the end. 3) Solutions can be generated at any desired time point and frequency. 

Commonly, even if the sensors are hardware synchronized, they still measure at distinct time points, causing the timestamps to be misaligned. Rather than interpolating \cite{niu2022ic} or rounding \cite{qin2019general} to align the timestamps, we create nodes in distinct timestamps for each sensor. The INS pre-integrations are then used to connect the nodes. This strategy is more effective because the INS time-propagation is more accurate than interpolating or rounding.

We should also consider the hardware delay of sensors in the real-time scenarios. The hardware delays differ between sensors, leading to instances where measurements with earlier timestamps are received after the processing of measurements with more recent timestamps. In such situations, the new coming state may be inserted or prepended to the graph. One solution is to introduce a delay in order to re-sequence the arriving measurements. However, it will lead to a loss of real-time performance. Instead, our choice is to apply inserting or prepending by re-establishing states using INS pre-integration. All the sensors can be added timely to ensure real-time performance.

Furthermore, we facilitate the integration of backend solutions at any timestamps and frequencies, within the capabilities of the INS, instead of limiting the output solely to camera or GNSS timestamps \cite{cao2022gvins, niu2022ic, qin2019general}. This approach will provide advantageous for downstream algorithmic modules and in particular, control algorithms, as it guarantees both strict real-time performance and bandwidth for navigation solutions.

By defining the factor graph and concretizing the graph as the LSQ problem in Equation \ref{eq:lsq}, the problem can be solved by any non-linear optimization algorithm. In GICI-LIB, we use Ceres-Solver \cite{Agarwal_Ceres_Solver_2022} for solving such LSQ problem. To concretize the graph, we should define the residual and Jacobian computation of the edges, i.e. factors. Hence, we will discuss the factors in the following.

\subsection{GNSS Factors}

The GNSS factors are classified into two levels: LC factors and TC factors. The LC factors mainly contain position factor and velocity factor, and the TC factors mainly contain pseudorange factor, doppler factor, and carrier-phase factor. Since the LC factors are simple, we only describe the TC factors here. 

The non-linear models of pseudorange, carrier-phase, and doppler measurements can be described as follows 
\begin{equation}
\label{eq:pseudorange}
P_{r, i}^{s}=\rho_{r}^{s}+c\left(d t_{r}-d t^{s}\right)+I_{r, i}^{s}+T_{r}^{s}+d_{r, i}-d_{i}^{s}+\varepsilon_{P}
\end{equation}
\begin{equation}
\label{eq:phaserange}
L_{r, i}^{s}=\rho_{r}^{s}+c\left(d t_{r}-d t^{s}\right)+I_{r, i}^{s}+T_{r}^{s}+b_{r, i}-b_{i}^{s}+N_{r, i}^{s}+\varepsilon_{L}
\end{equation}
\begin{equation}
\label{eq:doppler}
D_{r, i}^{s}=\dot{\rho}_r^s+c\left(d f_{r}-d f^{s}\right)+\varepsilon_{D}
\end{equation}
where indices $s$, $r$, and $i$ refer to the satellite, receiver, and carrier frequency band, respectively. $\rho_{r}^{s}=\left\|p_{r}-p^{s}\right\|^2$is the geometric distance between the receiver antenna phase center and satellite phase center. $ c $ is the speed of light in vacuum. $d t_{r}$ and $d t^{s}$ represent receiver clock offset and satellite clock offset respectively. $d f_{r}$ and $d f^{s}$ are the corresponding frequency offsets. $I_{r, i}^{s}$ is the ionospheric delay. $T_{r}^{s}$ is the tropospheric delay. $d_{r, i}$ and $ d^s_{i} $ are code biases for receiver and satellite. $b_{r, i}$ and $ b^s_{i} $ are phase biases. $N_{r, i}^{s}$ is the phase ambiguity, which has an integer nature. $\varepsilon_{P}$, $\varepsilon_{L}$, $\varepsilon_{D}$ are the un-modeled errors, mainly containing multipath and random noise.

\begin{table}[t]
\centering
\caption{Amplitude before and after correction (by model or service) for GNSS error sources in units of meters. For simplicity, we set the amp. as 0 if it is less than 1 mm.} 
\setlength{\tabcolsep}{1mm}
\renewcommand\arraystretch{1.2}
\belowrulesep=0pt
\aboverulesep=0pt
\begin{tabular}{c|l|l|l} 
    \toprule 
    {} & {Error source} & {Amp.} & {Amp. cor.} \\ 
    \hline
    \multirow{9}*{\makecell[c]{Space \\ segment}} & {Orbit error} & {-} & {0.01 $\sim$ 5} \\ 
    \cline{2-4}
    ~ & {Clock offset} & {-} & {0.01 $\sim$ 5} \\ 
    \cline{2-4}
    ~ & {Code bias} & {1 $\sim$ 10} & {0 $\sim$ 0.1} \\ 
    \cline{2-4}
    ~ & {Phase bias} & {0.01 $\sim$ 0.1} & {0 $\sim$ 0.02} \\ 
    \cline{2-4}
    ~ & {Phase center offset} & {1 $\sim$ 5} & {0 $\sim$ 0.01} \\ 
    \cline{2-4}
    ~ & {Phase center variation} & {0 $\sim$ 0.02} & {0} \\ 
    \cline{2-4}
    ~ & {Phase wind-up} & {0 $\sim$ 0.05} & {0} \\ 
    \cline{2-4}
    ~ & {GPS inter-frequency clock bias} & {0 $\sim$ 0.2} & {0 $\sim$ 0.02} \\ 
    \cline{2-4}
    ~ & {BDS satellite multipath} & {0.1 $\sim$ 0.8} & {0 $\sim$ 0.1} \\ 
    \hline
    \multirow{4}*{\makecell[c]{Propagation \\ segment}} & {Sagnac effect} & {$10^2 \sim 10^3$} & {0} \\ 
    \cline{2-4}
    ~ & {Ionosphere delay} & {5 $\sim$ 30} & {1 $\sim$ 3}  \\ 
    \cline{2-4}
    ~ & {Troposphere delay} & {1 $\sim$ 5} & {0.2 $\sim$ 1} \\ 
    \cline{2-4}
    ~ & {Relativistic effect} & {0 $\sim$ 0.02} & {0} \\
    \hline
    \multirow{9}*{\makecell[c]{Ground \\ segment}} & {Clock offset} & {0 $\sim$ $10^2$} & {-} \\ 
    \cline{2-4}
    ~ & {Code bias} & {1 $\sim$ 10} & {-} \\ 
    \cline{2-4}
    ~ & {Phase bias} & {0.01 $\sim$ 0.1} & {-} \\ 
    \cline{2-4}
    ~ & {Multipath} & {0 $\sim$ $10^2$} & {-} \\ 
    \cline{2-4}
    ~ & {Random noise} & {0.01 $\sim$ 0.1} & {-} \\ 
    \cline{2-4}
    ~ & {Phase center offset} & {0.01 $\sim$ 5} & {0 $\sim$ 0.01} \\ 
    \cline{2-4}
    ~ & {Phase center variation} & {0 $\sim$ 0.02} & {0} \\ 
    \cline{2-4}
    ~ & {Earth tide} & {0 $\sim$ 0.1} & {0 $\sim$ 0.01} \\ 
    \cline{2-4}
    ~ & {GLONASS inter-frequency bias} & {0.1 $\sim$ 2} & {0 $\sim$ 0.1} \\ 
    \bottomrule 
\end{tabular} 
\label{tab:gnss_errors}
\end{table}

There are also some other error sources that should be specially considered in addition to the above. We list all the error sources in Table \ref{tab:gnss_errors}. The ways to handle these errors vary among algorithms. In Equation \ref{eq:pseudorange}, \ref{eq:phaserange}, and \ref{eq:doppler}, we only maintain the items that may be estimated in GNSS formulations, presuming that all other errors have been handled appropriately.

As shown in Equation \ref{eq:pseudorange} and \ref{eq:phaserange}, there are still several error items maintained in the pseudorange and carrier phase equations. In GICI-LIB, we implemented 12 different formulations for each measurement type to account for the diverse characteristics of various GNSS algorithms. The formulations are combinations of two frame definitions, i.e. the Earth-Centered and Earth-Fixed (ECEF) frame or the East-North-Up (ENU) frame, three GNSS linear combination types, i.e. Zero-Difference (ZD), Single-Difference (SD), or Double Difference (DD), and two approaches handling atmosphere delays, i.e. estimated or non-estimate (corrected or eliminated). Table \ref{tab:gnss} summarizes those models and names the corresponding algorithms implemented in GICI-LIB.

\begin{table}[htbp]
\centering
\caption{Formulations of pseudorange and carrier-phase measurements towards the optimization problem. } 
\setlength{\tabcolsep}{1mm}
\renewcommand\arraystretch{1.2}
\belowrulesep=0pt
\aboverulesep=0pt
\begin{tabular}{p{1.5cm}<{\centering} | p{0.6cm}<{\centering} p{0.6cm}<{\centering} p{0.6cm}<{\centering} | p{0.9cm}<{\centering} p{0.9cm}<{\centering} | p{0.9cm}<{\centering} p{0.9cm}<{\centering}} 
    \toprule 
    {} & \multicolumn{3}{c|}{Differential} & \multicolumn{2}{c|}{Atmosphere} & \multicolumn{2}{c}{Frame} \\ 
    {} & {ZD} & {SD} & {DD} & {Cor./Eli.} & {Est.} & {ECEF} & {ENU} \\
    \hline 
    {SPP} & {P} & {} & {} & {P} & {} & {P} & {} \\
    {RTD} & {} & {} & {P} & {P} & {} & {P} & {} \\ 
    {RTK} & {} & {} & {P\&L} & {P\&L} & {} & {P\&L} & {} \\
    {PPP} & {P\&L} & {} & {} & {} & {P\&L} & {P\&L} & {} \\
    {SPP TC} & {P} & {} & {} & {P} & {} & {} & {P} \\
    {RTK TC} & {} & {} & {P\&L} & {P\&L} & {} & {} & {P\&L} \\
    \bottomrule 
\end{tabular} 
\label{tab:gnss}
\end{table}

The doppler measurement in Equation \ref{eq:doppler} exhibits less error than other measurements. Therefore, its differential or atmosphere handling is unnecessary. There are only two formulations implemented for doppler, categorized by their frames of reference. The doppler measurements are used in all the algorithms in Table \ref{tab:gnss}.

The above categories are formed towards the optimization problem, which are formulated by types of the estimated parameters and measurement models. To clarify, we only focus on the formulation of (short-baseline) RTK and PPP in the following. Detailed illustrations of all the formulations can be found in sections 3.3 and 3.4 of our documentation.

\subsection{Real-Time Kinermatic}

The measurements of RTK are DD pseudorange, DD carrier-phase, and ZD doppler. The DD operation differences measurements between GNSS receivers and satellites. For short-baseline RTK, we assume all the atmosphere delays are eliminated by DD \cite{takasu2009development}. After correcting all the necessary errors, the DD pseudorange and carrier-phase residuals can be written as
\begin{equation}
\label{eq:dd_pseudorange}
r_P = P_{r r_b,i}^{s s_b} - \rho_{r r_b}^{s s_b}
\end{equation}
\begin{equation}
\label{eq:dd_phaserange}
r_L = L_{r r_b,i}^{s s_b} - ( \rho_{r r_b}^{s s_b}+(\boldsymbol{N}_{r r_b, i}^{s}-\boldsymbol{N}_{r r_b, i}^{s_b}) )
\end{equation}
\begin{equation}
\label{eq:dd_doppler}
r_D = D_{r, i}^{s} - ( \dot{\rho}_r^s+c\left(d f_{r}-d f^{s}\right) )
\end{equation}
where $P_{r r_b,i}^{s s_b}$ and $L_{r r_b,i}^{s s_b}$ are the DD pseudorange and carrier-phase measurement respectively. $\rho_{r r_b}^{s s_b}$ is the DD geometric distance. $\boldsymbol{N}_{r r_b, i}^{s}$ is the SD ambiguity. We do not write the DD ambiguity here because we estimate the SD ambiguity instead to avoid having to frequently handle base satellite switching due to signal disruption from shadowing or dropouts.

Consequently, the GNSS-related parameters at epoch $k$ are defined as
\begin{equation}
\label{eq:rtk_parameters}
\boldsymbol{\chi}_{k} := \left[ \boldsymbol{\chi}_{I,k}, \boldsymbol{\chi}_{r,k} \right]^T
\end{equation}
\begin{equation}
\label{eq:rtk_parameters_ins}
\boldsymbol{\chi}_{I,k} := \left[{}^W \boldsymbol{t}_{I, k}, \boldsymbol{q}^W_{I, k}, {}^W \boldsymbol{v}_{I, k}  \right]^T
\end{equation}
\begin{equation}
\label{eq:rtk_parameters_gnss}
\boldsymbol{\chi}_{r,k} := \left[{}^B \boldsymbol{t}_r, d\boldsymbol{f}_{r,k}, \boldsymbol{N}_{r r_b, i, k}^{s} \right]^T
\end{equation}
where ${}^W \boldsymbol{t}_{I}$, $\boldsymbol{q}^W_I$, and ${}^W \boldsymbol{v}_I$ are the position, orientation, and velocity of INS respectively,  ${}^B \boldsymbol{t}_r$ is the translation of the GNSS receiver in the body frame $B$, namely lever-arm. $d\boldsymbol{f}_{r,k}$ is a vector of frequency offsets $d{f}_{r,k}$ for different satellite constellations. $\boldsymbol{N}_{r r_b, k}^{s}$ is a vector of SD ambiguities ${N}_{r r_b, k}^{s}$ for different satellites $s$ and frequencies $i$.

The above formulation can be used to form the RTK-only algorithm and then feed its solution into GNSS loose integration algorithms, and can also be used directly to form the GNSS tight integration algorithms. In comparison, the latter can utilize the propagated parameters assisted by other sensors to apply outlier detection on GNSS raw measurements. This feature makes it easier to reject problematic satellite signals and fully leverage the information provided by qualified signals.

\subsection{Precise Point Positioning}

The PPP algorithm employs undifferenced formulations, i.e. Equation \ref{eq:pseudorange}, \ref{eq:phaserange}, and \ref{eq:doppler}, to deliver positioning solutions at centimeter-level accuracy, without requiring local reference stations. Unlike RTK, the PPP algorithm should carefully handle several error items because it cannot apply the between-station difference to eliminate errors.

By applying SSR-I corrections, the satellite positions $p^s$, satellite clocks $dt^s$, and satellite code biases $d^s_i$ are corrected to centimeter-level accuracy, while still retaining a bias for each satellite system. The bias exists due to the coupling between the clock and the code bias, preventing the estimation of their absolute values. This unobservability also exists on the receiver side, due to the coupling between the receiver clock and bias. In brief, all the remaining code biases will be estimated or absorbed into the estimated parameters during estimation. One can find the detailed discussion on the bias aspect in section 3.4 of our documentation. 

For a multi-constellation and multi-frequency PPP, the estimated parameters, despite the INS parameters defined in Equation \ref{eq:rtk_parameters_ins}, are as follows:
\begin{equation}
\label{eq:ppp_parameters}
\boldsymbol{\chi}_{r, k} := \left[ {}^B \boldsymbol{t}_r, d\boldsymbol{t}_{r,k}, d\boldsymbol{f}_{r,k}, \boldsymbol{N}_{r, i, k}^s, T_{Z, w, k}, \boldsymbol{I}_{r, 1, k}^s, \boldsymbol{d}_{r,{IFB}_i} \right]^T  
\end{equation}

where $d\boldsymbol{t}_{r,k}$ and $d\boldsymbol{f}_{r,k}$ are vectors of receiver clocks and frequencies $d{t}_{r,k}$ and $d{f}_{r,k}$ for each system. $\boldsymbol{N}_{r, i, k}^s$ is a vector of ZD ambiguities $N_{r, i, k}^s$. $T_{Z, w, k}$ is the zenith total delay of the wet component of the troposphere. $\boldsymbol{I}_{r, 1, k}^s$ is a vector of ionosphere delays $I_{r, 1, k}^s$. $\boldsymbol{d}_{r,{IFB}_i}$ is a vector of inter-frequency biases ${d}_{r,{IFB}_i}$, which should be estimated for each frequency if the number of used frequencies is larger than two. Note that the aforementioned code biases will be absorbed into $d\boldsymbol{t}_{r,k}$, $\boldsymbol{I}_{r, 1, k}^s$, and ${d}_{r,{IFB}_i}$, rendering these parameters biased.

As we have not yet addressed the phase biases in Equation \ref{eq:phaserange}, the estimated ambiguities $N_{r, i, k}^s$ absorb these bias values, leading to the loss of their integer nature and cannot be used to solve integer ambiguities. In response, the SSR-II correction can be utilized to correct the satellite phase bias $b^s_i$. Meanwhile the receiver phase bias $b_{r, i}$ can be eliminated by forming a Between-Satellite-Difference (BSD). Then the ambiguity resolution algorithm can be applied.

However, both PPP and PPP-AR suffer from long convergence periods, typically lasting several or tens of minutes. To speed up their convergence, the SSR-III, containing local atmosphere delay corrections, could be applied as further constraints. Since this feature relies on local reference station networks, just as RTK, it is named PPP-RTK. The convergence period depends on the scale of reference station networks, ranging from tens of minutes to several seconds.

Currently, worldwide PPP services are underdeveloped and some of the standards have not been unified. Due to the absence of standardized and stable public products, we only support the SSR-I stream I/O in the current version. 

\subsection{GNSS Ambiguity Resolution}
\label{sec:Methodology:AR}

Ambiguity Resolution (AR) is a major procedure for high-precision GNSS algorithms to achieve centimeter-level solutions. This procedure is implemented for RTK and PPP (as well as the corresponding integration algorithms). We illustrate the RTK AR here. 

The AR starts with the estimated float ambiguities $\boldsymbol{N}_{r r_b, i, k}^{s}$ and their covariance matrix $\boldsymbol{P}_{N_{SD}}$. A BSD is first applied by selecting base satellites for each satellite constellation to eliminate the receiver phase biases $b_{r, i}$ absorbed by float ambiguities during estimation. Herein we get the DD ambiguities $\boldsymbol{N}_{r r_b, i, k}^{s s_b}$ and covariance matrix $\boldsymbol{P}_{N_{DD}}$. For convenience, we denote the DD ambiguities and their covariance as $\boldsymbol{N}$ and $\boldsymbol{P}$. Then the problem becomes solving an integer LSQ problem
\begin{equation}
\hat{\boldsymbol{N}} = \mathop{\arg\min}\limits_{\hat{\boldsymbol{N}}} \| \hat{\boldsymbol{N}} - \boldsymbol{N} \|^2_P
\label{eq:int_lsq}
\end{equation}
where $\hat{\boldsymbol{N}}$ is the integer ambiguities to be solved.

The problem can be solved by two steps: decorrelation and search. We use the MLAMBDA \cite{chang2005mlambda} algorithm to solve the problem. Once the integer ambiguities have been resolved and judged as valid resolutions, they will be constrained to FGO using pseudo-measurements with low variances.

Moreover, we apply partial ambiguity resolution and (ultra-) wide-lane combination technologies to further improve the fixation rate and reliability. The two technologies are essential to ensure AR performance, but have not been implemented in existing FGO-based GNSS-relevant algorithms.

\section{Experiments and Results}
\label{sec:Experiments}

We conducted multiple land vehicle experiments and employ public datasets, UrbanNav \cite{hsu2023hong}, to evaluate our PPP, RTK, and RTK-based integration algorithms, together with SOTA open-source algorithms with centimeter precision level, including RTKLIB \cite{takasu2009development}, OB-GINS \cite{tang2022impact}, IC-GVINS \cite{niu2022ic} and VINS-Fusion \cite{qin2019general}. In this section, we first introduce the experiment configurations of the datasets. Then the evaluation results are presented. Finally, an extra experiment is conducted to examine whether the performance is degraded when applying our flexible FGO structure. 

\begin{figure} [b]
    \centering
    \includegraphics[width=0.48\textwidth]{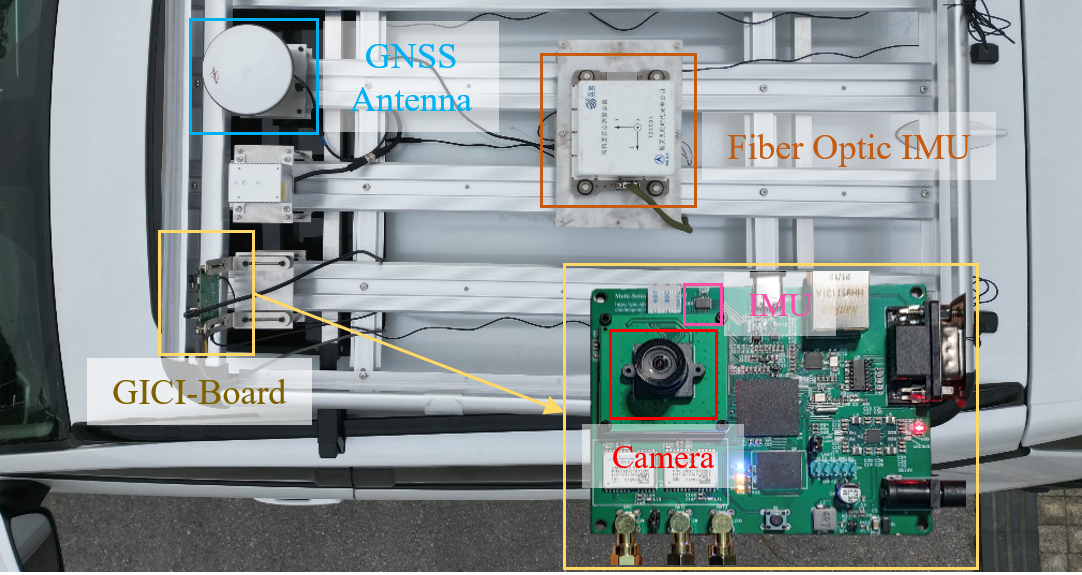}
    \caption{Experiment configuration of land vehicle experiments. }
    \label{fig:experiment_platform}
    \vspace{0.5em}
\end{figure}

\begin{figure} [htbp]
    \vspace{1em}
    \centering
    \includegraphics[width=0.48\textwidth]{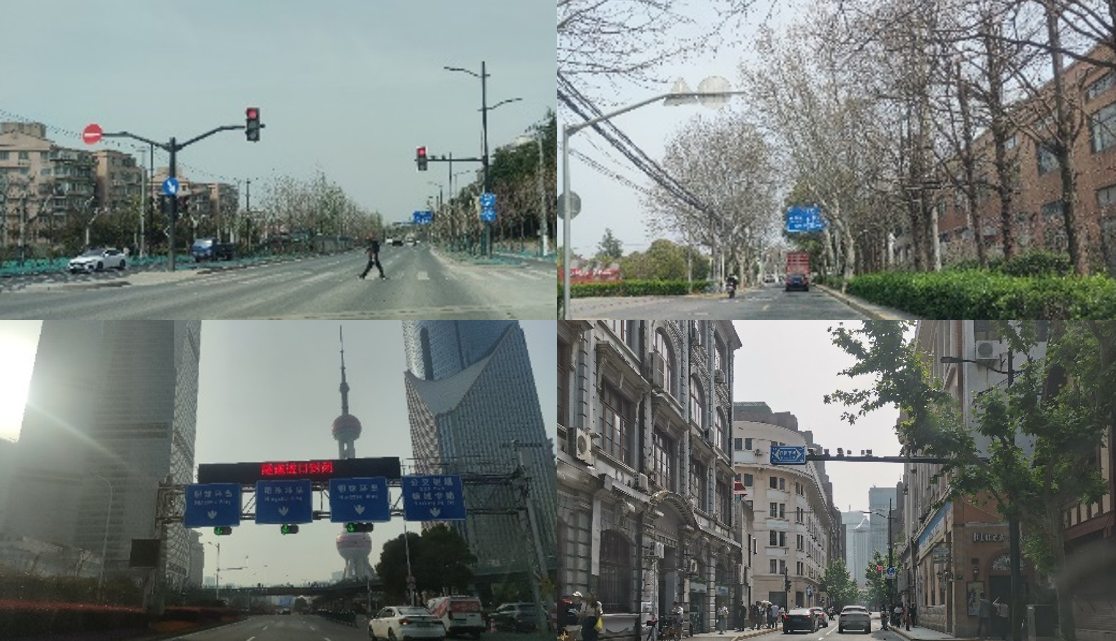}
    \caption{Type of scenes in short-term experiments. Top left, top right, bottom left, and bottom right are open-sky, tree-lined, typical urban, and dense urban, respectively.}
    \label{fig:scenes}
\end{figure}

\begin{table}[htbp]
\centering
\caption{GNSS properties of the scenes.} 
\setlength{\tabcolsep}{1mm}
\renewcommand\arraystretch{1.2}
\belowrulesep=0pt
\aboverulesep=0pt
\begin{tabular}{p{2.2cm} | p{1.0cm}<{\centering} p{1.0cm}<{\centering} p{1.0cm}<{\centering} | p{1.0cm}<{\centering} p{1.0cm}<{\centering}} 
    \toprule 
    {} & \multicolumn{3}{c|}{SPP Performance} & \multicolumn{2}{c}{Satellite Number}  \\ 
    {} & {Avail.} & {RMSE} & {Max} & {Mean} & {Max} \\
    \hline 
    {Open-sky} & {1.00} & {1.57} & {5.75} & {36} & {44} \\
    {Tree-lined} & {1.00} & {4.04} & {18.1} & {32} & {42} \\
    {Typical Urban} & {1.00} & {8.12} & {205.4} & {22} & {35} \\
    {Dense Urban} & {0.99} & {18.2} & {746.2} & {21} & {37} \\
    \hline 
    {UrbanNav mid} & {1.0} & {30.2} & {209.0} & {13} & {21} \\
    {UrbanNav deep} & {0.93} & {20.7} & {442.8} & {12} & {20} \\
    {UrbanNav harsh} & {0.99} & {39.3} & {472.1} & {13} & {18} \\
    \bottomrule 
\end{tabular} 
\label{tab:scene_spp}
\end{table}

\subsection{Experiment Configurations}

The configuration of our dataset is shown in Figure \ref{fig:experiment_platform}. We developed a GICI board to collect IMU and camera data and applied hardware synchronization with other sensors in the whole platform. The onboard IMU and camera are Bosch BMI088 and Onsemi MT9V034 respectively. The GNSS receiver is a Tersus David30 multi-frequency receiver. We also collected the reference station data from the Qianxun SI stream for RTD and RTK formulations, and the State-Space-Representation (SSR) data from the International GNSS Service (IGS) stream for PPP formulations. The fiber optic IMU is used to provide the ground truth by post-processing its data together with GNSS raw data.

We collected two kinds of datasets: short-term (3-10 minutes) experiments in different scenes, and long-term (20-50 minutes) experiments to cover different durations of considered challenging environments. For the short-term experiments, we categorize the scenes into 4 types: Open-sky, tree-lined, typical urban, and dense urban. And for each scene, we recorded 2 $\sim$ 3 trajectories. The scenes are illustrated in Figure \ref{fig:scenes}. For the long-term experiments, we provide two trajectories collected in the city center of Shanghai that cover those 4 types of scenes.

\begin{table*}[htbp]
\centering
\caption{Absolute Pose Error (APE) for each fusion algorithm in units of meters or meters/degrees. The minimum APEs in each dataset are bolded. '-' means the estimator diverged or the APE is larger than 100 m/deg.} 
\setlength{\tabcolsep}{1mm}
\setlength{\smallskipamount}{2pt}
\renewcommand\arraystretch{1.2}
\belowrulesep=0pt
\aboverulesep=0pt
\begin{tabular}{p{1.05cm} p{0.6cm}<{\centering} | p{0.65cm}<{\centering} p{0.65cm}<{\centering} p{0.65cm}<{\centering} | p{0.65cm}<{\centering} p{0.65cm}<{\centering} | p{1.4cm}<{\centering} p{1.4cm}<{\centering} p{1.4cm}<{\centering} p{1.4cm}<{\centering} | p{1.4cm}<{\centering} p{1.4cm}<{\centering} p{1.4cm}<{\centering}} 
    \toprule 
    \multicolumn{2}{c|}{} & \multicolumn{5}{c|}{GNSS-only Algorithms} & \multicolumn{7}{c}{Multi-sensor Fusion Algorithms} \\ 
    \cline{3-14}
    \multicolumn{2}{c|}{} & \multicolumn{3}{c|}{GICI-LIB} & \multicolumn{2}{c|}{RTKLIB} & \multicolumn{4}{c|}{GICI-LIB} & \multicolumn{3}{c}{SOTA} \\
    \multicolumn{2}{c|}{Dataset ID} & {RTK} & {DPPP} & {MPPP} & {RTK} & {DPPP} & {RTK LC} & {RTK TC} & {RTK SRR} & {RTK RRR} & {OB-GINS} & {IC-GVINS} & {VINS-F} \\ 
    \hline
    \multirow{2}*{Open-sky} & 
    {1.1} & {0.05} & {0.81} & {0.38} & {0.40} & {0.70} & {0.03 / 0.56} & {0.03 / 0.55} & {0.03 / 0.54} & {\textbf{0.03} / \textbf{0.54}} & {0.05 / 2.46} & {0.48 / 2.23} & {0.36 / 3.99} \\ 
    ~ & {1.2} & {\textbf{0.02}} & {0.35} & {0.38} & {0.07} & {0.35} & {0.03 / 1.70} & {0.03 / 1.69} & {0.04 / 1.74} & {0.04 / 1.69} & {0.09 / 4.54} & {0.04 / \textbf{0.88}} & {0.37 / 10.7} \\ 
    \hline 
    \multirow{2}*{Tree-lined} & 
    {2.1} & {0.19} & {2.09} & {1.25} & {0.85} & {5.52} & {0.19 / 0.57} & {0.16 / 0.58} & {0.19 / 0.56} & {\textbf{0.16} / \textbf{0.56}} & {- / -} & {1.32 / 1.60} & {0.41 / 8.73} \\ 
    ~ & {2.2} & {2.58} & {3.62} & {2.54} & {0.96} & {4.55} & {2.57 / 1.51} & {\textbf{0.27} / 1.53} & {2.57 / 1.49} & {0.31 / \textbf{1.39}} & {2.52 / 7.07} & {2.58 / 2.22} & {2.61 / 15.9} \\ 
    \hline 
    \multirow{3}*{\makecell[l]{Typical \\ Urban}} & 
    {3.1} & {0.88} & {1.92} & {1.62} & {2.07} & {6.40} & {0.87 / \textbf{1.49}} & {\textbf{0.16} / 1.60} & {0.87 / 1.56} & {0.29 / 1.58} & {56.8 / 26.7} & {32.0 / 16.4} & {17.6 / 34.1} \\ 
    ~ & {3.2} & {0.81} & {6.46} & {2.64} & {10.2} & {21.2} & {0.73 / 1.45} & {0.19 / 1.47} & {0.71 / 1.39} & {\textbf{0.19} / \textbf{1.26}} & {- / -} & {72.3 / 2.36} & {2.65 / 23.8} \\ 
    ~ & {3.3} & {0.60} & {3.38} & {2.31} & {1.57} & {4.48} & {0.28 / 1.45} & {\textbf{0.28} / 1.45} & {0.29 / 1.43} & {0.30 / 1.28} & {12.9 / 4.12} & {0.60 / \textbf{0.90}} & {- / -} \\ 
    \hline 
    \multirow{3}*{\makecell[l]{Dense \\ Urban}} & 
    {4.1} & {0.11} & {3.94} & {3.07} & {3.07} & {6.27} & {0.10 / 0.52} & {\textbf{0.07} / 0.52} & {0.10 / \textbf{0.51}} & {0.08 / 0.54} & {0.14 / 1.56} & {0.22 / 0.68} & {0.37 / 33.9} \\ 
    ~ & {4.2} & {6.45} & {4.82} & {2.96} & {10.2} & {8.07} & {10.7 / 1.99} & {2.45 / 1.45} & {10.1 / 1.64} & {\textbf{2.09} / \textbf{1.39}} & {55.1 / 97.5} & {13.8 / 4.30} & {- / -} \\ 
    ~ & {4.3} & {-} & {8.59} & {16.8} & {33.0} & {28.0} & {88.6 / 3.05} & {2.19 / \textbf{1.75}} & {83.7 / 2.90} & {\textbf{1.61} / 1.76} & {- / -} & {- / -} & {- / -} \\ 
    \hline 
    \multirow{2}*{\makecell[l]{Long \\ Term}} & 
    {5.1} & {1.89} & {2.44} & {1.65} & {3.78} & {5.62} & {1.87 / 0.68} & {0.48 / \textbf{0.60}} & {1.28 / 0.63} & {\textbf{0.21} / 0.63} & {5.09 / 3.01} & {3.79 / 1.13} & {- / -} \\ 
    ~ & {5.2} & {2.46} & {2.38} & {3.59} & {2.71} & {3.99} & {2.40 / 0.52} & {0.14 / 0.53} & {2.39 / \textbf{0.43}} & {\textbf{0.11} / 0.49} & {33.6 / 2.16} & {2.96 / 0.91} & {- / -} \\ 
    \hline 
    \multicolumn{2}{l|}{UrbanNav mid} & {6.69} & {/} & {/} & {7.84} & {/} & {6.70 / 3.31} & {\textbf{2.24} / \textbf{1.17}} & {8.36 / 1.98} & {3.40 / 1.30} & {- / -} & {48.1 / 7.75} & {14.6 / 13.2} \\
    \multicolumn{2}{l|}{UrbanNav deep} & {8.74} & {/} & {/} & {13.8} & {/} & {8.85 / 3.16} & {3.93 / 1.88} & {7.81 / 2.14} & {\textbf{2.46} / \textbf{1.64}} & {- / -} & {54.1 / 11.0} & {- / -} \\ 
    \multicolumn{2}{l|}{UrbanNav harsh} & {21.0} & {/} & {/} & {33.2} & {/} & {18.0 / 5.14} & {7.04 / 3.55} & {21.3 / 2.65} & {\textbf{6.73} / \textbf{1.44}} & {- / -} & {- / -} & {- / -} \\
    \bottomrule 
\end{tabular} 
\label{tab:fusion_ape}
\end{table*}

Furthermore, we employ the UrbanNav dataset \cite{hsu2023hong} for a more comprehensive evaluation. We utilize a subset of its sensor configurations, including a Ublox F9P double frequency receiver, an Xsens-MTI-30 IMU, and the left camera of the ZED module. Additionally, we utilize the Hong Kong Satellite Positioning Reference Station Network (SatRef) as the reference station for the RTK algorithm. Since there are no SSR messages recorded, we can not apply the PPP algorithm for this dataset.

To demonstrate the criteria for selecting scenarios, we present the GICI-SPP results and the number of satellites in Table \ref{tab:scene_spp}. Here, the availability represents the percentage of valid solutions over the whole period with an expected solution output rate (10 Hz for the GICI dataset and 1 Hz for UrbanNav). Please note that the data in Table \ref{tab:scene_spp} is influenced by both the scenes and the quality of GNSS equipment.

\subsection{Performance Evaluation}

We evaluate the Absolute Pose Error (APE) for each trajectory by comparing the real-time outputs of the algorithms with the ground truth. The results are summarized in Table \ref{tab:fusion_ape}. The PPP algorithms are divided into dual-frequency PPP (DPPP) and multi-frequency PPP (MPPP) to ensure a fair comparison because the RTKLIB only supports DPPP. Since OB-GINS, IC-GVINS, and VINS-Fusion rely on GNSS solutions as inputs, their workflows are fed using the solution produced by GICI-LIB RTK. Moreover, the initialization procedure of IC-GVINS often fails under non-fixed GNSS statuses, so we adjusted the starting point of the trajectories for IC-GVINS to ensure valid initialization. Besides, OB-GINS lacks an initialization algorithm, thus we use the initialization output of IC-GVINS for OB-GINS as they are part of the same software series.

It is clear that our algorithms outperform the other open-source software in most of the datasets. For the RTK algorithms, GICI-LIB leverages the advantage of FGO in better handling the non-linearity, compared with the EKF-based RTKLIB. Furthermore, the improved AR algorithm, as described in Section \ref{sec:Methodology:AR}, enhances the performance of GICI-RTK by increasing the fixation rate and reliability, thereby enabling the attainment of more centimeter-level epochs. In terms of PPP algorithms, GICI-PPP outperforms RTKLIB due to the use of the FGO algorithm, improved handling of various GNSS error sources, and support for multiple frequencies.

The multi-sensor fusion algorithms also outperform other SOTA algorithms. The result of the VINS-Fusion is expected owing to its loose integration paradigm. Although OB-GINS and IC-GVINS have the same formulation as our LC and SRR algorithms, they still failed to perform as expected. The major reason is that they only utilized GNSS position, but did not use GNSS velocity, which makes their estimator having difficulty in handling the significant nonlinearity. There are also some other minor helpers for our better performance, such as more comprehensive motion constraints and better outlier detection.

Focusing on the GICI-LIB algorithms, we notice that the estimators achieving greater accuracy are those using GNSS tight integrations. This is because raw GNSS measurements are utilized, which enables easier identification of outliers for each satellite, and permits satellite measurements to be used when the number of observed satellites is insufficient for RTK-only estimation. Moreover, the RRR estimator reaches the best performance in most of the trajectories due to the exhaustive utilization of multi-sensor measurements, but the improvement is not as much as one can expect relative to the TC estimator because of our greater tendency towards GNSS when we design the entire estimator. One can explore the ability of visual estimation to further improve the performance of the visual-relevant estimators.

\subsection{Validation of the Flexible FGO Structure}

In this section, we validate whether the performance is degraded when applying our flexible FGO structure to arbitrarily add sensor measurements. The GICI RTK SRR algorithm is used to conduct several strategies of measurement addition. The strategies are: S1) Re-sequencing measurements with a 0.2 s time buffer, and rounding the GNSS timestamps to nearby image timestamps, as similar as \cite{qin2019general}. S2) Re-sequencing with the same configuration, and interpolating GNSS timestamps to nearby image timestamps, as similar as \cite{niu2022ic}. S3) Re-sequencing and creating nodes in distinct timestamps. S4) Flexible inserting and creating nodes in distinct timestamps, which is utilized in GICI-LIB. To ensure real-time performance, the output timestamp is aligned to the latest timestamp of IMU measurement. Every time when a new IMU measurement is received, the latest backend solution will be integrated to the IMU timestamp and outputted.

\begin{table}[t]
\centering
\caption{Absolute Pose Error (APE) for the four strategies of FGO construction in units of meters/degrees.} 
\setlength{\tabcolsep}{1mm}
\renewcommand\arraystretch{1.2}
\belowrulesep=0pt
\aboverulesep=0pt
\begin{tabular}{p{1.5cm}<{\centering} | p{1.4cm}<{\centering} p{1.4cm}<{\centering} p{1.4cm}<{\centering} p{1.4cm}<{\centering}} 
    \toprule 
    {Dataset ID} & {S1} & {S2} & {S3} & {S4}  \\ 
    \hline 
    {5.1} & {1.46 / 0.64} & {1.34 / 0.65} & {1.42 / \textbf{0.61}} & {\textbf{1.28} / 0.63} \\
    {5.2} & {2.64 / 0.44} & {2.57 / 0.45} & {2.54 / 0.46} & {\textbf{2.39} / \textbf{0.43}} \\
    \bottomrule 
\end{tabular} 
\label{tab:validate_flexibility}
\end{table}

We revisit our long-term datasets to conduct these experiments. The APEs of the four strategies are shown in Table \ref{tab:validate_flexibility}. The results show that our proposed strategy achieves the highest performance. Strategies S1 and S2 perform unsatisfactorily due to the loss of measurement precision when applying rounding or interpolating. Strategy S3 does not outperform due to the longer INS integration duration for output, caused by the re-sequencing buffer. 

Consequently, our FGO structure is able to guarantee high precision whilst ensuring user flexibility.

\section{Conclusion}
\label{sec:Conclusion}

In this letter, we proposed a GNSS/INS/Camera integrated navigation library and released a comprehensive dataset to fully leverage the potential of GNSS algorithms. The integration scheme and precision grade of GNSS formulations are clarified. By evaluating the GNSS-only and RTK-based integration algorithms, we demonstrated that GNSS can exhibit an extraordinary and crucial role in providing precise fused solutions if the capabilities of its measurements are exhaustively exploited. While our implementations were based on the fundamental GNSS formulations, we anticipate that exploring the relationships between these formulations and improving outlier rejection strategies will further enhance the performances. We hope this work could assist the community to further explore the potential of GNSS-based multi-sensor integrated navigation.


%



\ifCLASSOPTIONcaptionsoff
  \newpage
\fi



%



\bibliographystyle{IEEEtran}
\bibliography{gici.bib}

%





\end{document}